\documentclass[10pt]{article}
\usepackage[a4paper, total={5in, 7in}]{geometry}
\usepackage[usenames,dvipsnames,svgnames,table]{xcolor} 
\usepackage{authblk}

\usepackage{algorithm}
\usepackage{algorithmic}
\floatname{algorithm}{Algorithm}

\usepackage[
  backend=biber,
  style=nature,
  sorting=none,
  maxbibnames=5
]{biblatex}
\DeclareFieldFormat{doi}{\url{https://doi.org/#1}}
\usepackage{longtable}
\usepackage{float}
\newfloat{algorithm}{tb}{lop}
\usepackage{amsmath, amssymb, amsfonts, amsthm}
\usepackage{enumitem} 
\usepackage{graphicx}
\usepackage{tikz}
\usetikzlibrary{3d, calc}
\usepackage{bm} 
\usepackage{hyperref}
\usepackage{multirow}
\usepackage{subcaption}
\usepackage{array}
\newcolumntype{L}{>{\centering\arraybackslash}m{0.1\linewidth}}

\newtheorem{definition}{Definition}

\newcommand{\Exp}{\textrm{Exp}}
\newcommand{\Log}{\textrm{Log}}

\newcommand{\Diag}{\mathrm{Diag}}
\newcommand{\Off}{\mathrm{Off}}
\newcommand{\diagvec}{\operatorname{diag}}
\newcommand{\Cor}{\mathrm{Cor}^+}
\newcommand{\Hol}{\mathrm{Hol}}
\newcommand{\Expoff}{\operatorname{Exp}_{\mathrm{off}}}
\newcommand{\Logoff}{\operatorname{Log}_{\mathrm{off}}}
\newcommand{\iso}{\cong}

\newcommand{\tr}{\operatorname{tr}}

\begin{document}
	
\title{Riemannian geometry meets fMRI: the advantages of modeling correlation manifolds and eigenvector subspaces}
\author[1,*]{Mario Severino}
\author[1,2]{Manuela Moretto}
\author[3,4,5]{Robert A. McCutcheon}
\author[1,2]{Mattia Veronese}

\affil[1]{Department of Information Engineering, University of Padova, Padova, Italy}
\affil[2]{Department of Neuroimaging, Institute of Psychiatry, Psychology and Neuroscience (IoPPN), King’s College London, London, United Kingdom}
\affil[3]{Department of Psychiatry, University of Oxford, Warneford Hospital, Warneford Ln, Headington, Oxford OX3 7JX, United Kingdom}
\affil[4]{Oxford Health NHS Foundation Trust, Warneford Hospital, Warneford Ln, Headington, Oxford OX3 7JX, United Kingdom}
\affil[5]{Department of Psychosis Studies, Institute of Psychiatry, Psychology and Neuroscience, King’s College London, De Crespigny Park, London, SE5 8AF, United Kingdom}

\affil[*]{email: mario.severino@phd.unipd.it}

\date{}

\maketitle
\begin{abstract}

Correlation matrices are fundamental summaries of functional brain networks, yet standard analyses often treat entries independently, ignoring the curved geometry of correlation space. Existing geometric methods frequently lack closed-form operations or depend on arbitrary region ordering, limiting scalability.
We introduce a scalable geometric framework with two components: (i) the Off–log metric, a smooth transformation mapping correlation matrices to symmetric zero-diagonal matrices. This enables closed-form expressions for distances, Fréchet means, and linear models, allowing standard statistical modeling without complex manifold optimization. (ii) Grassmannian subspace discrimination, which compares subjects via principal-angle distances between eigenvector subspaces, resolving inherent sign and basis ambiguities.
Both components integrate into standard machine-learning workflows for inference, regression, and classification. Validated across two clinical cohorts (Parkinson’s and psychosis) and three ageing fMRI datasets, the Off––log metric increased sensitivity in permutation tests and matched or exceeded Riemannian and Euclidean baselines in classification. Brain-age prediction performance was comparable, with  Riemannian metrics excelling in two of three cohorts. The Grassmannian method consistently outperformed Euclidean baselines, highlighting disease-relevant networks. Overall, geometry-aware representations improve sensitivity and predictive performance while remaining straightforward to deploy at scale.

\end{abstract}


\section{Introduction}\label{sec:intro}

Over the past two decades, a rapidly expanding body of work has pushed machine learning and data analysis beyond the conventional Euclidean setting to incorporate the richer mathematical structures of non-Euclidean geometry, topology, and abstract algebra \cite{Papillon2025-gs, Snasel2017-wo}. This broader movement includes extensions of classical statistical theory and inference, often grouped under \textit{geometric statistics} \cite{Guigui2023-gj, Pennec2006-ga}, as well as recent advances in deep learning that explicitly leverage geometric, topological, and equivariant principles \cite{Bronstein2021-mw, Bodnar2022-ty, Hajij2022-jf, Cohen2021-zn}.

In many domains, data possess an inherent spatial or structural organization, for example, the spatial layout of a brain scan or the surface geometry of a protein. Even datasets that lack an obvious spatial embedding can often be interpreted as samples drawn from an underlying low-dimensional manifold immersed in a high-dimensional ambient space \cite{Whiteley2022-ez}. Characterizing the ``shape'' of this underlying space---the geometry that governs the data---frequently uncovers meaningful relationships and patterns that improve interpretation and downstream analysis.

\begin{figure}[tbp] 
  \centering
  \includegraphics[width=0.8\linewidth]{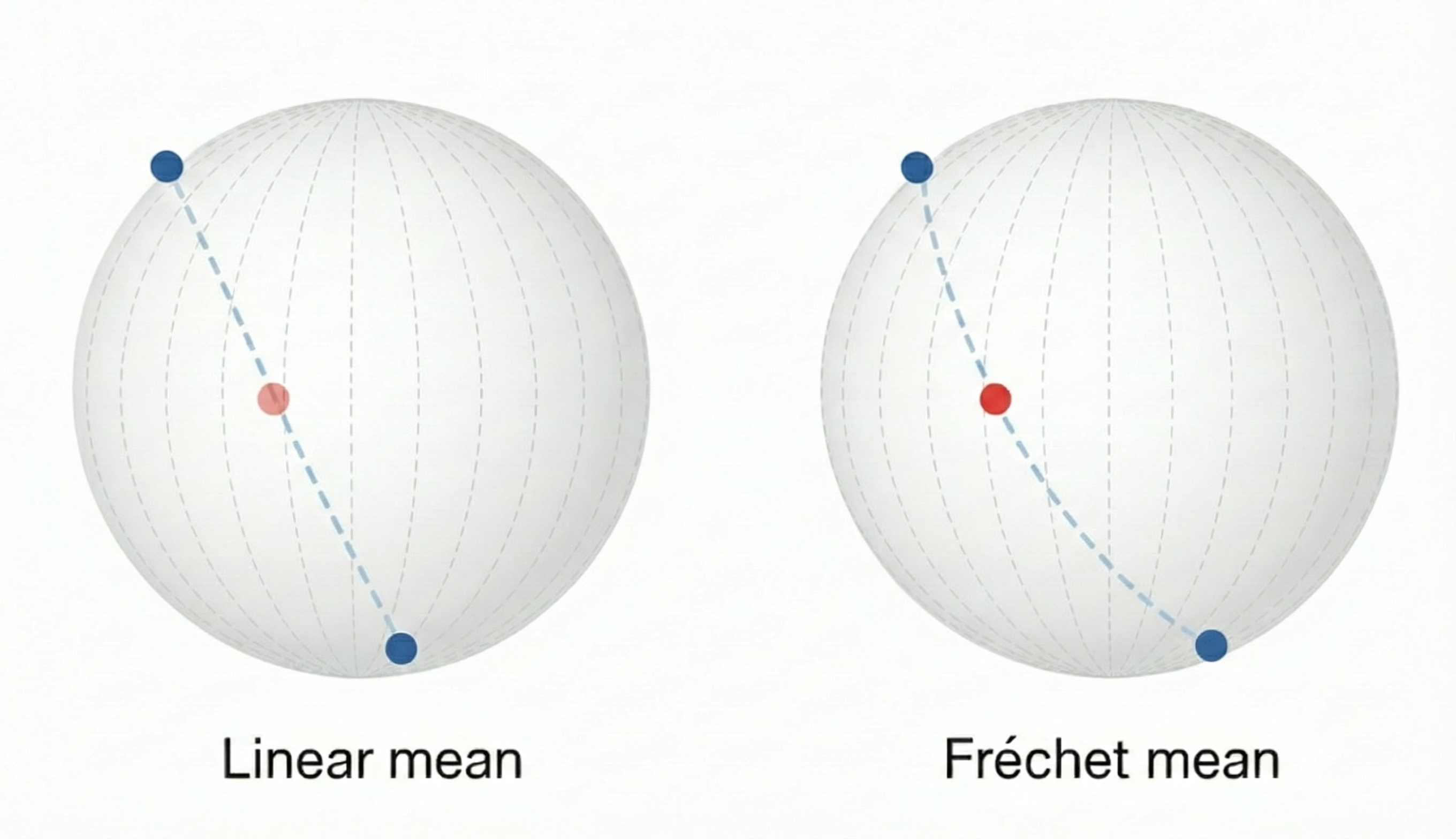} 
  \caption{Left: The Euclidean (linear) mean follows the ambient-space chord between two points on the sphere and typically falls off the surface into the interior of the sphere, which does not belong to the manifold.
Right: The Fréchet (Riemannian) mean minimizes the sum of squared geodesic distances and lies at the geodesic midpoint on the sphere.}
  \label{fig:mean}
\end{figure}

In neuroscience, the brain is commonly modeled as a network of interacting regions, and these interactions are often summarized using second-order statistics such as covariance or correlation estimated from resting-state functional MRI (rs-fMRI) \cite{Van_den_Heuvel2010-em}, positron emission tomography (PET) \cite{Severino2025-co}, electroencephalography (EEG), and magnetoencephalography (MEG) recordings \cite{Brookes2011-pc, Cohen2014-ab}. Brain network analyses based on correlation matrices frequently ignore the dependency structure among matrix entries; yet, when considered as whole objects, correlation matrices encode substantially more information than the mere set of pairwise correlations. Consequently, it is important to treat correlation matrices as manifold-valued data endowed with an appropriate geometric structure. For example, the simple Euclidean average of a set of correlation matrices generally does not lie on the manifold that contains the data, whereas the geometric (Fréchet) mean provides a more faithful representation of the intrinsic ``center of mass'' of the data (Figure.~\ref{fig:mean}).

Full--rank correlation matrices (hereafter ``correlation matrices'') belong to the broader family of symmetric positive definite (SPD) matrices, which also includes covariance matrices, and this family carries a natural Riemannian manifold structure \cite{Thanwerdas-undated-na}. In practice, working directly with correlation matrices on the SPD manifold can be inconvenient: routine matrix operations may produce matrices that no longer have a unit diagonal, which then requires additional normalization to recover a valid correlation structure \cite{Thanwerdas-undated-na}.

In contrast to the SPD manifold, the natural mathematical space tailored for correlation matrices is the elliptope \cite{Thanwerdas-undated-na}. Despite being the proper setting, the elliptope has historically received relatively little attention, largely because its intrinsic metric, the quotient-affine metric (QAM), lacks a closed-form expression, which makes it challenging to apply in high-dimensional settings such as neuroimaging \cite{Thanwerdas-undated-zn}. In recent years, however, pioneering studies have introduced alternative geometric frameworks on the elliptope, enabling closed-form solutions by exploiting diffeomorphisms (maps that preserve smooth structure) to Euclidean space \cite{Thanwerdas2022-iv}. Some of these newly developed metrics have already been applied to neuroimaging data (fMRI or EEG), yielding promising results \cite{You2022-vj, You2025-ta}.

Another geometric object that has received comparatively little attention in neuroscience, despite its clear relevance when eigenvectors are used, is the Grassmannian manifold \cite{Bendokat2020-fu}. This viewpoint is natural whenever data analysis relies on eigen decompositions: for example, the principal subspace spanned by the first $k$ eigenvectors of a correlation matrix is an element of the Grassmannian. Treating these eigenspaces as points on a manifold, rather than as ordered lists of basis vectors, removes ambiguities due to eigenvector sign flips or orthonormal basis rotations and focuses on the subspace geometry that is truly meaningful. In neuroimaging, Grassmannian-based methods can be applied wherever eigenspaces are informative: comparing principal connectivity modes across subjects, summarizing dominant spatial patterns from principal component analysis (PCA) \cite{Nguyen2022-bc, Fan2011-lz}, independent component analysis (ICA) \cite{Selvan2012-sh}, or the harmonics (eigenvectors) of the graph Laplacian.

The primary aim of this work is to introduce Riemannian–geometric methods into neuroimaging and to show that they provide a more mathematically principled treatment of correlation-based data, which in turn can improve statistical learning performance on both small clinical cohorts and large open datasets. Specifically, we study the Off--log Euclidean Riemannian metric on the manifold of correlation matrices. The Off--log metric yields a permutation-invariant, log-Euclidean–type geometry tailored to correlation matrices. We compare this geometry with previously used correlation-manifold metrics and develop a Grassmannian discriminant method that operates on the subspaces spanned by eigenvectors of the graph Laplacian constructed from fMRI-derived correlation matrices, taken here as a representative instance of correlation-based neuroimaging data.

The paper is organized as follows. First, we review the necessary elements of Riemannian geometry for correlation data analysis. Second, we present the Off--log metric on the correlation manifold and summarize the relevant Grassmannian theory used for subspace discrimination. Third, we analyze the proposed tools, both computationally and theoretically, and validate the full pipeline on clinical neuroimaging datasets of varying sizes from resting-state fMRI data.

\section{Background}\label{sec:background}

\subsection{Basics of Riemannian geometry}

A \textit{manifold} is a topological space that, in the neighborhood of every point, looks like Euclidean space (i.e., it is locally homeomorphic to $\mathbb{R}^{n}$). In plain terms: locally the manifold is linear and non-self-intersecting, but its global shape may curve. Crucially, a manifold is not a vector space, and-operations taken for granted in Euclidean settings (addition, subtraction, scalar multiplication) are not defined globally. From a data-analysis viewpoint this is important because standard statistical and machine-learning methods assume vector-space structure and therefore cannot be applied directly to manifold-valued data.

One natural way to introduce distances and angles on a manifold is to endow it with a \textit{Riemannian metric}. Intuitively, a Riemannian metric is a smoothly varying, positive-definite inner product on tangent vectors that plays the role of a ``local ruler'' at each point. Practically, the metric tells you how to measure the length of an infinitesimal displacement at a given location. Using this metric, the geodesic distance (i.e., distance in curved geometry) between two points on the manifold is defined as the length of the shortest curve connecting them, obtained by integrating these infinitesimal lengths along the curve.

\begin{figure}[!t] 
  \centering
  \includegraphics[width=0.55\textwidth]{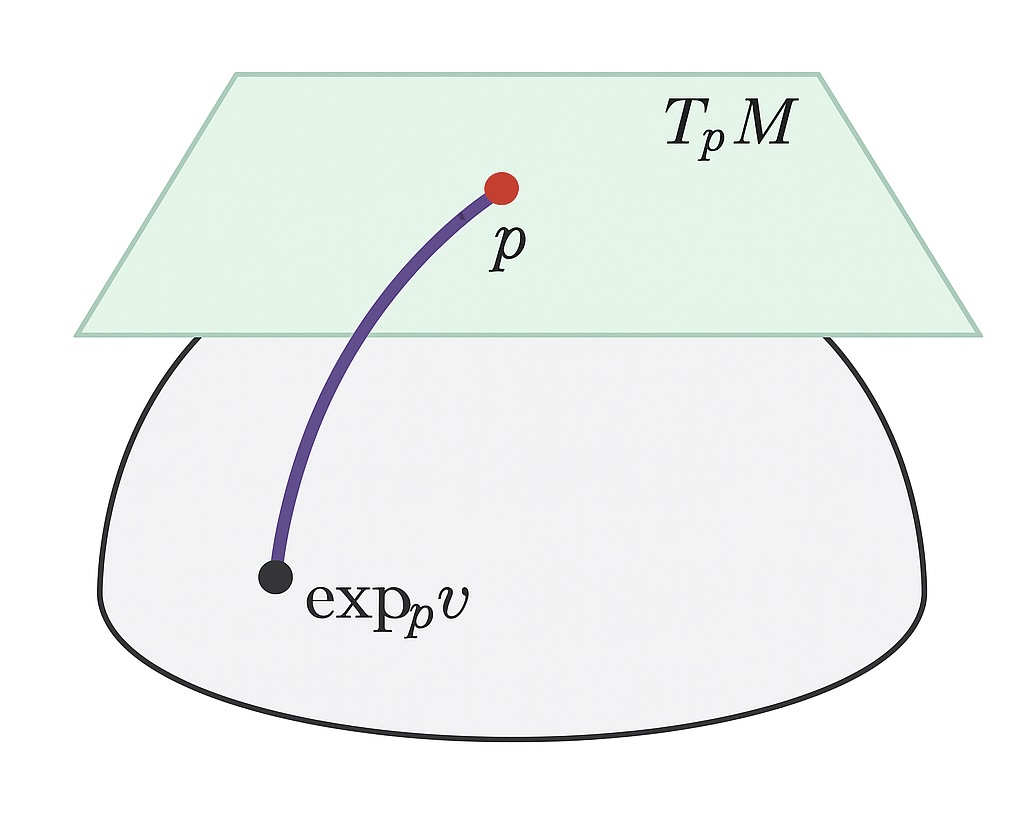} 
  \caption{At any point $p$, the tangent space $T_pM$ provides the local linear approximation of the manifold. The exponential map $\exp_p$ takes a direction and step in $T_pM$ and returns the point on the manifold reached by traveling along the corresponding geodesic from $p$.}
  \label{fig:exponential}
\end{figure}

At a given base point on a manifold, the \textit{tangent space} is the best linear approximation to the manifold near that point. It can be pictured as a flat $n$-dimensional plane that just touches the curved surface at the point of interest. For example, at a point on a sphere the tangent space is the plane that touches the sphere at that point. A tangent vector points in some direction along the surface and indicates the initial direction of a great-circle (Figure.~\ref{fig:exponential}). In computations, the tangent space is central because it enables a three-step strategy: (i) map manifold points to a tangent space (via a logarithm or chart) where Euclidean methods apply, (ii) perform statistical or learning operations there, and (iii) map results back to the manifold (via the exponential map). This local linearization is the standard way to adapt vector-space algorithms to manifold-valued data. The choice of the base point at which the linearization is carried out can significantly influence the outcome; common choices include the identity element or the Fréchet mean.

Another algebraic notion with a deep connection to Riemannian manifolds is that of a \textit{group}, defined as a set of elements equipped with a rule for combining them (a binary operation). This operation must satisfy certain axioms \cite{Canals2012-nb}. Groups can be discrete or continuous. A \textit{Lie group} is a continuous group, defined as a smooth manifold equipped with a compatible group structure such that the composition of elements on the manifold obeys the group axioms \cite{Canals2012-nb}. Because of this dual nature, every point on a Lie group has both a geometric and an algebraic meaning. In particular, at the identity element of a Lie group one can look at its tangent space, which forms what is called the \textit{Lie algebra} of the group (a vector space). The Lie algebra can be thought of as capturing the ``infinitesimal directions'' in which one can move away from the identity along the manifold, and its elements can be seen as generators of smooth curves on the group. This connection between the Lie group and its Lie algebra provides a powerful way to move between nonlinear and linear settings in computations on manifold-valued data.

\subsection{Geometry of SPD and correlation manifolds}

Brain functional connectivity is typically summarized with second-order statistics such as covariance and correlation matrices. These objects lie in the class of SPD matrices, whose space is formally defined as follows:

\begin{definition} 
	$\mathrm{Sym}^+(n)$ is the space of $(n \times n)$ symmetric positive-definite matrices: 
	\begin{equation*} 
		\ \mathrm{Sym}^+(n)= \big\{ X \in \mathbb{R}^{n \times n} \mid X = X^\top,~ \lambda_{\min}(X) > 0 \big\}, 
	\end{equation*} 
	where $\lambda_{\min}(\cdot)$ denotes the smallest eigenvalue of the matrix. 
\end{definition}

The natural Riemannian metric on $\mathrm{Sym}^+(n)$ is the affine-invariant Riemannian metric (AIRM), which allows one to measure distances on this manifold \cite{Thanwerdas2023-mw}. This is not the unique metric that can be defined on $\mathrm{Sym}^+(n)$ \cite{Arsigny2007-sk}. In this work we concentrate on functional connectivity represented by correlation matrices, a constrained subset of SPD matrices. The space of correlation matrices, denoted $\mathrm{Cor}^+(n)$, is formally defined as follows:

\begin{definition} 
	$\mathrm{Cor}^+(n)$ is the space of $(n \times n)$ symmetric positive-definite matrices with unit diagonal elements: 
	\begin{equation*} 
		\ \mathrm{Cor}^+(n) = \big\{ X \in \mathbb{R}^{n \times n} \mid X \in \ \mathrm{Sym}^+(n),~ \diagvec(X) = \boldsymbol{1}_n \big\}, 
	\end{equation*} 
	where $\diagvec(X)$ is the vector of diagonal elements of $X$, and $\boldsymbol{1}_n$ is the vector of ones. 
\end{definition}

This definition establishes that $\mathrm{Cor}^+(n)$ is a strict subset of $\mathrm{Sym}^+(n)$ \cite{Thanwerdas-undated-na}. Indeed, a correlation matrix is obtained from a covariance matrix by normalizing each entry by the product of the corresponding marginal standard deviations. Equivalently, forming the diagonal matrix of standard deviations and then pre and post multiplying the covariance matrix by its inverse yields the correlation matrix. Consequently, any two covariance matrices that differ only by positive diagonal rescaling of the variables, that is, are diagonally congruent, induce the same correlation matrix. 
More generally, the space of correlation matrices can be endowed with a Riemannian structure via the theory of quotient manifolds. In particular, \cite{Thanwerdas-undated-zn} introduced the QAM metric, which inherits AIRM from $\mathrm{Sym}^+(n)$ and therefore provides a principled Riemannian framework for correlation matrices.
While QAM offers a rigorous geometric treatment of $\mathrm{Cor}^+(n)$, it becomes computationally demanding as either the number of connectivity matrices grows or the dimensionality of the variables-of-interest space increases. Consequently, recent theoretical work has sought to define simpler geometric structures on $\mathrm{Cor}^+(n)$ that retain useful geometric properties while reducing computational cost \cite{Thanwerdas2022-iv}.
For further details on the AIRM and QAM geometries in the context of brain functional connectivity analysis, we direct readers to earlier works \cite{You2022-vj, You2025-ta}.

\subsection{Pullback metrics on the correlation manifold}

A pullback metric transfers a Riemannian metric from one manifold to another (diffeomorphic) manifold via a smooth invertible map. Typically, a metric from a ``simpler'' Euclidean space is pulled back to the manifold of interest, allowing computations to be performed in Euclidean coordinates while preserving geometric structure on the manifold.
Among the geometries recently proposed for the space of correlation matrices, two constructions have gained widespread use because they provide closed-form, computationally tractable formulas while retaining many useful geometric properties.

The \emph{Euclidean--Cholesky metric} (ECM) maps a correlation matrix $C \in \mathrm{Cor}^+(n)$ to its Cholesky factor 
\[
L=\operatorname{chol}(C) \;\in\; \mathrm{LT}^1(n),
\]
belonging to the space of lower-triangular matrices with unit diagonal. The Cholesky factor ensures $C = LL^\top$. Since the Cholesky map is smooth, the Euclidean inner product on $\mathrm{LT}^1(n)$ can be pulled back to define a Riemannian metric on $\mathrm{Cor}^+(n)$. In practice, this reduces manifold operations such as distances, means, and interpolations to elementary vector operations on the Cholesky coordinates, which are fast and numerically stable \cite{Thanwerdas2022-iv}.

The \emph{Log--Euclidean Cholesky} (LEC) variant introduces an additional logarithmic reparametrization to better handle multiplicative constraints. It defines a composite diffeomorphism
\[
\mathrm{Cor}^+(n) \longrightarrow \mathrm{LT}^1(n) \longrightarrow \mathrm{LT}^0(n),\qquad
C \longmapsto L=\operatorname{chol}(C) \longmapsto \log L,
\]
where $\mathrm{LT}^0(n)$ denotes lower-triangular matrices with zero diagonal and $\log$ denotes the matrix logarithm. The standard Euclidean inner product on $\mathrm{LT}^0(n)$ is then pulled back to $\mathrm{Cor}^+(n)$, yielding a flat log-Euclidean geometry, where all computations can be performed in closed form in Euclidean coordinates \cite{Thanwerdas2022-iv, You2025-ta}.

Both ECM and LEC avoid costly operations by mapping all matrices to Euclidean coordinates, applying standard algorithms there, and mapping the results back to $\mathrm{Cor}^+(n)$. For detailed mathematical treatments we refer the reader to \cite{Thanwerdas2022-iv}.

A major drawback of these Cholesky-based constructions is their lack of permutation invariance: the Cholesky factorization depends on the ordering of variables, so permuting rows/columns changes the coordinate representation and hence the metric's results. In contrast, permutation-invariant metrics assign the same distance regardless of variable order, which is often desirable when no canonical ordering exists.

Because of the computational advantages of closed-form Euclidean pullbacks and the practical importance of permutation invariance, recent work has explored alternative mappings and metrics that aim to combine both properties \cite{Thanwerdas2024-ar}. In the next subsection we focus on one such proposal: the \emph{Off--log Euclidean metric}.

\subsection{Off--log Euclidean metric}

The \emph{Off--log Euclidean} metric was first introduced in \cite{Thanwerdas2024-ar} with the aim of endowing $\mathrm{Cor}^+(n)$ with a permutation-invariant log-Euclidean metric that can be computed in closed form and that possesses theoretically convenient properties (as opposed to QAM). 
The Off--log bijection $\Logoff : \mathrm{Cor}^+(n) \to \mathrm{Hol}(n)$ between the vector space of symmetric hollow matrices (zero diagonal) and correlation matrices is a smooth diffeomorphism. Since this map is equivariant under permutations, it enables the pullback of families of permutation-invariant inner products on hollow matrices to correlation matrices.

\begin{definition}\
An Off--log metric on $\mathrm{Cor}^+(n)$ is the pullback metric of a permutation-invariant inner product characterized by a quadratic form $q$ on $\mathrm{Hol}(n)$ via the diffeomorphism $\Logoff : \mathrm{Cor}^+(n) \to \mathrm{Hol}(n)$.
\end{definition}
In plain terms, this means that we can use a simplified Euclidean metric defined on $\mathrm{Hol}(n)$ to calculate distances on $\mathrm{Cor}^+(n)$.

The transformation is defined such that, given $C \in \mathrm{Cor}^+(n)$,
\[
\Logoff(C) := \Off(\log C) \;=\; \log C - \Diag\!\big(\diagvec(\log C)\big),
\]
where $\log$ denotes the matrix logarithm, $\diagvec$ the operator that extracts the diagonal of a matrix as a vector, and by $\Diag$ the operator that maps a vector to the corresponding diagonal matrix. For any $S\in\Hol(n)$ let $D(S)\in\Diag(n)$ be the (unique) diagonal matrix satisfying
\[
\diagvec\!\big(\exp\!\big(S + D(S)\big)\big) \;=\; \boldsymbol{1}_n,
\]
where $\exp$ is the matrix exponential, $\Diag(n)$ is the vector space of diagonal matrices and $\boldsymbol{1}_n$ is the vector of ones. Then, the inverse map is defined by
\[
\Expoff:\Hol(n)\longrightarrow \Cor(n),\qquad
\Expoff(S) := \exp\!\big(S + D(S)\big),
\]
where the diagonal correction $D(S)$ is chosen so that $\Expoff(S)\in\Cor(n)$. The existence and uniqueness of $D(S)$ are guaranteed in \cite{Archakov2020-vr, Johnson2009-pj}. Note that the Riemannian exponential and logarithm maps coincide with the diffeomorphisms $\Expoff : \mathrm{Hol}(n) \to \mathrm{Cor}^+(n)$ and $\Logoff : \mathrm{Cor}^+(n) \to \mathrm{Hol}(n)$ only at $ C = I_n$ \cite{Thanwerdas2024-ar}.

\paragraph{Off--log Euclidean Metric.}
Let \(\langle\cdot,\cdot\rangle_{\Hol}\) denote the canonical Frobenius inner product on \(\Hol(n)\),
\[
\langle A,B\rangle_{\Hol} := \tr(A^\top B),\qquad A,B\in\Hol(n).
\]
The Off--log Riemannian metric \(g_{\mathrm{off}}\) is the pullback of \(\langle\cdot,\cdot\rangle_{\Hol}\) by \(\Logoff\):
\[
g_{\mathrm{off}}\big|_{C}(U,V) \;=\; \big\langle d\Logoff|_C(U) ,\, d\Logoff|_C(V)\big\rangle_{\Hol},
\qquad U,V\in T_C\Cor,
\]
where $T_C\Cor$ is the tangent space of $\Cor$ at $C$.

Informally, the Off--log map lets us treat correlation matrices as points in a flat space of ``log-correlations'': we take a matrix logarithm, remove the diagonal, and work with the resulting hollow matrix as a vector. Distances, averages, and statistical models are then computed in this simple Euclidean setting, and the inverse map $\Expoff$ guarantees that the results are always brought back to valid correlation matrices. In practice, this provides a principled yet computationally convenient way to compare and average correlation patterns across subjects or experimental conditions.
Therefore, the Off--log diffeomorphism provides a closed-form distance between two correlation matrices, modulo the computation of a symmetric matrix logarithm.

\subsubsection*{Lie group structure of the Off--log diffeomorphism}

It has been shown that $\Cor(n)$ endowed with the Off--log metric has a \emph{log-Euclidean Lie group} structure \cite{Bisson-undated-nh}. The vector space $\Hol(n)$, with the addition of a binary operation $+$, is an abelian (i.e., commutative) Lie group. We transport this group law to $\Cor(n)$ via the diffeomorphism $\Expoff$.

\begin{definition}
Define a binary operation $\star$ on $\Cor(n)$ by pulling back the additive structure of $\Hol(n)$:
\[
C_1 \star C_2 \;:=\; \Expoff\big(\Logoff(C_1) + \Logoff(C_2)\big),\qquad C_1,C_2\in\Cor(n).
\]
Then $(\Cor(n),\star)$ is a smooth abelian log-Euclidean Lie group, and the inverse of $C\in\Cor(n)$ is
\[
C^{-1}_{\star} \;=\; \Expoff\big(-\Logoff(C)\big).
\]
Because such groups are simply connected and commutative, the Lie exponential is exactly $\Expoff$. The corresponding Lie algebra is the tangent space at the identity
\[
\mathfrak g \;=\; T_{I_n}\Cor \;=\; \Hol(n).
\]
Because the group law is induced by vector addition, the Lie bracket on \(\mathfrak g\) is trivial (the algebra is commutative), and the Lie exponential/logarithm coincide with the maps \(\Expoff:\mathfrak g\to\Cor\) and \(\Logoff:\Cor\to\mathfrak g\) introduced above.
\end{definition}

\paragraph{Metric at the identity and bi-invariance.}

The Off--log Riemannian metric \(g_{\mathrm{off}}\) evaluated at a point $C\in\Cor(n)$ was previously defined as
\[
g_{\mathrm{off}}\big|_{C}(U,V) \;=\; \big\langle d\Logoff|_C(U) ,\, d\Logoff|_C(V)\big\rangle_{\Hol},
\qquad U,V\in T_C\Cor.
\]
Evaluating it at the identity gives an immediate simplification: since \(d\log|_{I_n}=\mathrm{Id}\) and \(\Off\) acts trivially on hollow matrices, for \(U,V\in T_{I_n}\Cor=\Hol(n)\) we have
\[
d\Logoff|_{I_n}(U) = U,\qquad\text{hence}\qquad
g_{\mathrm{off}}\big|_{I_n}(U,V)=\tr(UV).
\]
Thus the metric at the identity coincides with the Frobenius inner product \(\langle\cdot,\cdot\rangle_{\Hol}\) on \(\Hol(n)\) $\iso$ Lie\((\Cor(n))\) (the Lie algebra of the \(\Cor(n)\) group). Moreover, because \(\Logoff\) is a Lie group isomorphism onto \((\Hol(n),+)\), left (and right) translation on \((\Cor,\star)\) corresponds to Euclidean translation in \(\Hol(n)\), and therefore the metric \(g_{\mathrm{off}}\) is \emph{translation invariant} (indeed bi-invariant in this abelian case): for every \(A\in\Cor\) and \(U,V\in T_C\Cor\),
\[
g_{\mathrm{off}}\big|_{A\star C}\big( dL_A(U),\, dL_A(V)\big)
\;=\; g_{\mathrm{off}}\big|_{C}(U,V),
\]
where \(L_A\) denotes left multiplication by \(A\) in the \(\star\)-product. In coordinates, this identity follows from the relation
\[
\Logoff(A\star C) \;=\; \Logoff(A) + \Logoff(C),
\]
so the derivative with respect to $C$ is unchanged by left translation.

From a practical viewpoint, the $\star$-operation means that combining or interpolating correlation matrices amounts to adding their Off--log representations and mapping back with $\Expoff$. Geodesics and averages in this Lie group therefore become straight lines and ordinary averages in Off--log coordinates, while still respecting the nonlinear constraints of correlation matrices. This log-Euclidean Lie group structure thus offers a conceptually clean and numerically stable way to model smooth changes in connectivity patterns.

\paragraph{Computational considerations.}
These algebraic and metric definitions have several immediate practical consequences for computation:

\begin{itemize}
\item \textbf{Working in the Lie algebra equals working in the tangent space at the identity.} Mapping \(C\mapsto S=\Logoff(C)\in\mathfrak g\) identifies every correlation matrix with a unique hollow symmetric matrix. Because \(\Expoff\) and \(\Logoff\) are (globally) inverse diffeomorphisms, operations performed linearly on \(S\) have an exact manifold meaning after mapping back via \(\Expoff\). In other words, in this setting the Lie algebra \(\mathfrak g\) provides global coordinates on \(\Cor\), not only a local linear approximation.
\item \textbf{Geodesics and distances are linear in the algebra.} For \(C_1,C_2\in\Cor\) with \(S_i=\Logoff(C_i)\), the geodesic and Riemannian distance are
  \[
  \gamma(t)=\Expoff\big((1-t)S_1+tS_2\big),\qquad
  d_{\mathrm{off}}(C_1,C_2)=\|S_2-S_1\|_F.
  \]
  Therefore, interpolation, means, and Fréchet averages reduce to arithmetic operations in \(\Hol(n)\):
  \begin{equation}\label{eq:log-frechet-mean}
  \overline C \;=\; \Expoff\!\Big(\frac{1}{m}\sum_{i=1}^m \Logoff(C_i)\Big).
  \end{equation}
\item \textbf{Group operations are cheap in the algebra.} A residual update \(S\mapsto S+\Delta S\) in \(\mathfrak g\) corresponds exactly to left (or right) multiplication by the group element \(\Expoff(\Delta S)\) on \(\Cor\):
  \[
  \Expoff(S+\Delta S) \;=\; \Expoff(\Delta S)\star \Expoff(S).
  \]
  Hence implementing manifold-native neural layers can be done by (i) computing Euclidean updates in the vectorized \(S\)-space and (ii) mapping back with \(\Expoff\) to obtain valid correlation matrices.
\item \textbf{The diagonal correction $D(S)$.} The only nontrivial computational step when mapping from \(\mathfrak g\) back to \(\Cor\) is the evaluation of
\[
\Expoff(S)=\exp\!\big(S + D(S)\big),
\]
because \(D(S)\) is defined implicitly by \(\diagvec\!\big(\exp(S+D(S))\big)=\boldsymbol{1}_n\). In practice one can compute \(D(S)\) by an iterative algorithm as described in \cite{Archakov2020-vr}.
\end{itemize}

We conclude this section by summarizing the advantages of adopting the proposed geometries. First, through diffeomorphic transformations, these geometries retain the fundamental characteristics of standard Euclidean space, effectively exhibiting zero curvature. This property allows the direct application of conventional statistical and machine-learning techniques without the additional complexity often associated with non-Euclidean domains. Moreover, it opens the possibility of designing neural network architectures that incorporate the group operations introduced above, thereby combining the computational simplicity of Euclidean methods with the theoretical rigor and structural consistency provided by the underlying geometric framework. Finally, although tangent–space projections around a base point are in general only local linear approximations, in our case the Off–log map itself is a global diffeomorphism between the correlation manifold and its Lie algebra. Thanks to the bi-invariance of the metric any point on the manifold can be transported to the identity (and back) by left or right translations, so projecting to the Lie algebra (the tangent space at the identity) gives a globally valid linear representation of all points.

\subsection{Grassmannian manifold}

When data analysis relies on eigenvectors, such as the leading principal components of a correlation matrix or the eigenvectors of a graph Laplacian, the resulting vectors are not uniquely defined. In particular, any eigenvector can be multiplied by $-1$ without altering the subspace it spans, and in the presence of repeated eigenvalues the associated eigenvectors can be arbitrarily rotated within their eigenspace. Consequently, two distinct orthonormal bases may correspond to the same geometric subspace, despite having different matrix representations. The Grassmannian formalism resolves this ambiguity by considering subspaces, rather than ordered bases, as the fundamental objects: all bases spanning the same subspace are identified with a single point on the Grassmannian \cite{Bendokat2020-fu}. This perspective eliminates spurious variability induced by sign flips, rotations within eigenspaces, or arbitrary basis orderings, thereby rendering Grassmannian-based distances and statistics intrinsically robust to such indeterminacies in spectral analyses.

\begin{definition}
The \emph{Grassmannian} $\mathrm{Gr}(k,n)$ is the set of all $k$-dimensional linear subspaces of $\mathbb{R}^n$:
\[
\mathrm{Gr}(k,n) \;=\; \{\, V \subset \mathbb{R}^n \;:\; V \text{ is a linear subspace and } \dim(V)=k \,\}.
\]
\end{definition}
It is the natural geometric object when the quantity of interest is a subspace rather than an ordered basis. One standard representation is via orthonormal basis matrices: every element of the Grassmannian may be written as the equivalence class $[U]$, where $U\in\mathbb{R}^{n\times k}$ satisfies $U^\top U = I_k$ and $[U]=\{UQ:Q\in O(k)\}$, with $O(k)$ the orthogonal group.

A fundamental notion for comparing two subspaces $[U]$ and $[V]$ is given by their \emph{principal angles} $\{\theta_i\}_{i=1}^k$. If $U$ and $V$ are orthonormal bases for the two subspaces, the singular value decomposition
\[
U^\top V \;=\; M \, \mathrm{diag}(\cos\theta_1,\dots,\cos\theta_k)\, N^\top,
\]
yields $\cos\theta_i$ as the $i$-th singular value of $U^\top V$ and $\theta_i=\arccos(\sigma_i)$. A common distance induced by these angles is the geodesic distance:
\[
d_{\mathrm{geo}}([U],[V]) \;=\; \left( \sum_{i=1}^k \theta_i^2 \right)^{1/2},
\]
Principal angles are used to compute Fréchet means on the Grassmannian:
\[
\overline{[U]} \;=\; \arg\min_{[X]\in\mathrm{Gr}(k,n)} \sum_{j} d_{\mathrm{geo}}^2([X],[U_j]),
\]
typically solved by alternating log/exp updates in the tangent space.

These algebraic and differential structures make the Grassmannian particularly useful in neuroimaging, where eigenspaces are routinely used: examples include comparing principal connectivity modes across subjects (PCA subspaces of covariance/correlation matrices), constructing subspace-based features for classification or regression \cite{Fan2011-lz}, and tracking time-varying subspaces in dynamic functional connectivity \cite{Dan2022-bp}. Practically, most computations reduce to robust linear-algebra kernels (SVDs and projector arithmetic), which scale well and avoid ambiguities due to basis rotations or sign indeterminacies of individual eigenvectors.


\section{Materials and methods}\label{sec:methods}

This section presents the methodological framework for analyzing (i) fMRI correlation matrices, modeled as points on the correlation manifold, and (ii) eigenvector matrices of the graph Laplacian built from those correlation matrices, modeled as points on the Grassmannian.

\subsection{Datasets}
\subsubsection{Healthy ageing datasets}
The first dataset  (hereafter \textit{cam-CAN}) was derived from the Cambridge Centre for Ageing and Neuroscience repository (\url{https://cam-can.mrc-cbu.cam.ac.uk/dataset/}). It consisted of MRI data from 627 healthy subjects (53.8$\pm$18.5 years, 312M/315F). MRI data were acquired at the Medical Research Council Cognition and Brain Sciences Unit in Cambridge on a 3T Siemens TIM Trio scanner equipped with a 32-channel head coil. The MRI protocol comprised a Magnetization-Prepared Rapid Gradient-Echo (MPRAGE) sequence (repetition time (TR) = 2250 ms; echo time (TE) = 2.99 ms; inversion time (TI) = 900 ms; flip angle (FA) = $9^{\circ}$; voxel size = $1 \times 1 \times 1$ mm$^3$), and around 9 minutes of rs-fMRI acquired with a single-band gradient-echo echo-planar imaging (TR = 1970 ms; TE = 30 ms; FA = $78^{\circ}$; voxel size = $3 \times 3 \times 4.4$ mm$^3$; number of volumes = 261). The second dataset (hereafter \textit{HCPAging}) was derived from the “Lifespan Human Connectome Project in Aging” (HCP-Aging) \cite{BOOKHEIMER2019335, HARMS2018972}. It consisted of MRI data from 599 healthy subjects (58.6$\pm$14.7 years, 262M/337F). MRI data were acquired on a 3T Siemens Prisma scanner equipped with a 32-channel head coil. The MRI protocol comprised a multi-echo MPRAGE sequence (TR = 2500 ms; TEs = 1.8/3.6/5.4/7.2 ms; TI = 1000 ms; FA = $8^{\circ}$; voxel size = 0.8 × 0.8 × 0.8 mm$^3$), and around 7 minutes of rs-fMRI acquired with a multi-band gradient-echo echo-planar imaging (TR = 870 ms; TE = 37 ms; FA = $52^{\circ}$; voxel size = 2 × 2 × 2 mm$^3$; number of volumes = 488; multi-band acceleration factor = 8). The third dataset (hereafter \textit{NKI}) was derived from the “Nathan Kline Institute Rockland Sample” (NKI-RS) \cite{Nooner2012-nk}. It consisted of MRI data from 894 healthy subjects (46.7$\pm$17.7 years, 303M/591F). MRI data were acquired on a 3T Siemens TIM Trio scanner equipped with a 32-channel head coil. The MRI protocol comprised a MPRAGE sequence (TR = 1900 ms; TE = 2.52 ms; TI = 900 ms; FA = $9^{\circ}$; voxel size = 1 × 1 × 1 mm$^3$), and around 10 minutes of rs-fMRI acquired with a multi-band gradient-echo echo-planar imaging (TR = 1400 ms; TE = 30 ms; FA = $65^{\circ}$; voxel size = 2 × 2 × 2 mm$^3$; number of volumes = 418; multi-band acceleration factor = 4).

\subsubsection{The Parkinson's disease dataset}
This dataset (hereafter \textit{DataPD}) consisted of MRI data from 20 patients with Parkinson’s disease (PD; 63.8$\pm$10.1 years) and 17 healthy controls (HC; 55.8$\pm$10.6 years). This dataset was acquired for the AND-PD project, a study carried out at the Parkinson’s Foundation Center of Excellence at King’s College Hospital, London. All subjects underwent a one-hour 3T MRI acquisition on a 3T GE scanner equipped with a 32-channel coil, consisting of functional and structural MRI sequences performed at the Center for Neuroimaging Sciences (CNS, KCL). The MRI protocol included the following sequences: an anatomical Fast Gray Matter Acquisition T1w Inversion Recovery (FGATIR) 3D sequence (TR = 6150 ms; TE = 2.1 ms; TI = 450 ms; FA = $8^{\circ}$; voxel size = $1 \times 1 \times 1$ mm$^3$), and 7 minutes of rs-fMRI acquired with multiband gradient-echo echo-planar imaging (TR = 890 ms; TE = 39 ms; FA = $50^{\circ}$; voxel size = $2.7 \times 2.7 \times 2.4$ mm$^3$; multiband acceleration factor = 6; number of volumes = 495).

\subsubsection{The Non-affective psychosis dataset}
This dataset (hereafter \textit{DataNAP}) consisted of MRI data from 60 patients with non-affective psychosis (NAP), onset within five years prior to study entry; diagnoses included schizophrenia, schizophreniform, schizoaffective disorder, psychosis NOS, delusional disorder, or brief psychotic disorder, and 57 HC. Participants were recruited as part of the Human Connectome Project (HCP) – Early Psychosis and scanned across four sites: Indiana University, Beth Israel Deaconess Medical Center–Massachusetts Mental Health Center, McLean Hospital, and Massachusetts General Hospital.
All MRI data were acquired on Siemens MAGNETOM Prisma 3T scanners equipped with either a 32-channel or 64-channel head coil, using a multiband acceleration factor of 8. Each subject underwent four runs of rs-fMRI collected across two experimental sessions on consecutive days (two runs per session). Resting-state fMRI was acquired with a multiband echo-planar imaging sequence (TR = 720 ms; voxel size = $2 \times 2 \times 2$ mm$^3$; number of volumes = 410 per run; scan duration = 4 min 55 s). Phase-encoding direction alternated between anterior–posterior (AP) and posterior–anterior (PA) across the two runs of each session. Structural T1-weighted images were also acquired following the HCP standard protocol. Detailed inclusion and exclusion criteria are available in~[insert reference].

\subsection{fMRI correlation matrices}

Following standard preprocessing (see Appendix~\ref{sec:appendix} for full details), the rs-fMRI data from the three ageing datasets were parcellated into 200 regions of interest according to the Schaefer seven-network atlas \cite{Schaefer2018-mf}. Similarly, \textit{DataPD} was parcellated into 400 regions, again based on the Schaefer seven-network atlas, whereas \textit{DataNAP} was parcellated into 360 regions defined by the Glasser atlas \cite{Glasser2016-vw}. The functional connectivity matrices were subsequently derived by computing Pearson’s correlations between the mean time series of the regions.

\subsection{Hypothesis testing on correlation matrices}
\label{subsec:exploratory-testing}

Let \(\{C_i\}_{i=1}^m\subset\Cor\) denote the set of subject-wise functional connectivity matrices, where \(\Cor\) is the manifold on which correlation matrices are modeled under a chosen geometry (ECM, LEC, Off--log). When observations take values on a nonlinear metric space, the appropriate notion of a sample average is the \emph{Fr\'echet mean}, which in its most commonly used form is defined by
\begin{equation}\label{eq:frechet-mean}
\overline C \;=\; \arg\min_{C\in\Cor}\; \frac{1}{m}\sum_{i=1}^m d^2(C,C_i),
\end{equation}
where \(d(\cdot,\cdot)\) denotes the geodesic distance induced by the chosen Riemannian metric.

Existence and uniqueness of the minimizer in \eqref{eq:frechet-mean} depend on the geometry. The ECM, LEC, and Off--log metrics ensure a unique Fréchet mean \cite{Thanwerdas2022-iv, Thanwerdas2024-ar}. In the case of the Off--log metric, \eqref{eq:frechet-mean} reduces to \eqref{eq:log-frechet-mean}, a simple arithmetic mean in log-Euclidean coordinates.

To enable standard multivariate inference while retaining geometric corrections, we employ the usual tangent-space linearization around the identity (which, for the Off--log metric, coincides with its Lie algebra). Each individual functional connectivity matrix is represented by its tangent vector at $I_n$:
\[
X_i \;=\; \Log_{I_n}(C_i)\;\in\; T_{I_n}\Cor,
\]
and, if desired, these tangent vectors can be vectorized to produce Euclidean feature vectors, which can then be used with classical statistical tests and machine-learning pipelines, as done in the subsequent analyses.

For hypothesis testing we adapted an interpoint-distance-based two-sample test in the spirit of Biswas and Ghosh \cite{Biswas2014-ux, You2025-ta}. Conceptually, the Biswas--Ghosh test compares average within-group dissimilarities to average between-group dissimilarities to form a test statistic that is sensitive to differences in the multivariate distribution of the two samples and that performs well in high-dimensional, low-sample regimes. A natural Biswas--Ghosh statistic is
\[
T_{\mathrm{BG}} \;=\; \overline{d}_{AB} \;-\; \tfrac{1}{2}\big(\overline{d}_{AA}+\overline{d}_{BB}\big),
\]
where $\overline{d}_{AB}$ is the between-group distance, while $\overline{d}_{AA}$ and $\overline{d}_{BB}$ are within-group distances. Larger values of \(T_{\mathrm{BG}}\) indicate greater separation between the two groups relative to within-group cohesion. The null distribution of \(T_{\mathrm{BG}}\) is obtained by permutation of group labels 1000 times (nonparametric permutation test), from which a \(p\)-value is computed. Thus, we applied this statistical test to raw correlation matrices and tangent-space-projected correlation matrices under the Off--log metric.

\subsection{Brain age prediction}
\label{subsec:brain-age}

We employed the three ageing datasets (\textit{cam-CAN, HCPAging and NKI}) to perform a brain-aging regression task using subject-wise fMRI functional connectivity matrices. The goal of this experiment was to assess whether accounting for the geometric structure of the data could improve predictive performance on well-established neuroimaging benchmarks. To this end, we trained an Elastic Net regression model to predict chronological age from connectivity patterns. We chose the Elastic Net because it provides a balance between the sparsity-promoting effect of $L_1$ regularization and the stability offered by $L_2$ regularization, making it well suited for high-dimensional and potentially collinear features such as functional connectivity measures. 

We evaluated two types of input representations: (i) the raw correlation matrices treated as Euclidean objects, and (ii) the same matrices after being mapped to the tangent space at the identity under the chosen geometry (ECM and Off--log). The tangent-space projection allows the data to be represented (in general) in a locally linear Euclidean space while preserving the manifold-informed structure. Each matrix was vectorized by extracting its upper-triangular entries, yielding feature vectors that were standardized and then reduced via PCA. To avoid data leakage, PCA was fitted using only the training subjects within each cross-validation fold and subsequently applied to the corresponding held-out subjects. The resulting low-dimensional representations were used as inputs to an Elastic Net regression model trained with nested 5-fold cross-validation. Model performance on unseen data was quantified on the outer held-out folds using mean absolute error (MAE) and the coefficient of determination ($R^2$).

\subsection{Classification}

We employed \textit{DataPD} and \textit{DataNAP} to address two distinct classification tasks: the first aimed at discriminating HC from individuals with PD, and the second at discriminating HC from individuals with NAP. A linear support vector machine (SVM) classifier was adopted to ensure model interpretability and to facilitate a clearer comparison between input representations. Specifically, we evaluated two types of features: the raw correlation matrices and their tangent-space projections under the chosen geometry (ECM, LEC, Off--log). We trained the classifier on the upper-triangular entries of the connectivity matrices using a nested cross-validation framework. For each fold, all preprocessing steps were fitted exclusively on the training data: univariate feature selection (ANOVA F-test), $z$-score standardization, and SVM training were implemented as a single pipeline and then applied to the corresponding held-out subjects. Hyperparameters were tuned via a 5-fold stratified inner cross-validation using the area under the receiver operating characteristic curve (AUC-ROC) as the optimization criterion, while generalization performance was assessed on the outer held-out folds using 5-fold stratified cross-validation. Classification performance was quantified using accuracy, AUC-ROC, sensitivity (recall of the disease class), and specificity (recall of the control class).

\subsection{Grassmannian discriminant analysis}

In this task we used \textit{DataPD} and \textit{DataNAP} to address the same two classification problems (HC vs.\ PD and HC vs.\ NAP) using subspaces spanned by graph Laplacian eigenvectors. For each subject, we first constructed a weighted adjacency matrix \(W\) from the fMRI correlation matrix by setting negative edges to zero and sparsifying weak connections using a 20\% threshold \cite{Luppi2024-yy}. We then computed the normalized graph Laplacian
\[
L \;=\; I - D^{-1/2} W D^{-1/2},
\]
where \(D=\mathrm{diag}(d_1,\dots,d_n)\) is the degree matrix with entries \(d_i=\sum_j W_{ij}\). The normalized Laplacian is symmetric positive semidefinite with eigenvalues in \([0,2]\). Its eigendecomposition
\[
L = U\Lambda U^\top,\qquad 
\Lambda=\mathrm{diag}(\lambda_1,\dots,\lambda_n),\quad
0=\lambda_1\leq \cdots\leq\lambda_n,
\]
yields orthonormal eigenvectors in the columns of \(U\). Eigenvectors associated with small eigenvalues correspond to low-frequency graph harmonics that vary smoothly over the graph and capture global connectivity structure, whereas eigenvectors with larger eigenvalues encode higher-frequency, more localized (and often noisier) variation \cite{Sipes2024-qs, Atasoy2016-mi, Belkin2003-db, Guo2018-jg}.

To select the dimensionality \(k\) of the low-frequency subspace, we used the \emph{gap spectrum}
\[
g_j \;=\; \lambda_{j+1} - \lambda_{j}, \qquad j=1,\dots,n-1,
\]
and chose \(k\) based on the largest eigenvalue gap, i.e., the index immediately after the most prominent separation between successive eigenvalues. A large gap indicates that the first \(j\) eigenvectors form a cohesive low-frequency subspace that is well separated from higher-frequency modes. Operationally, we first computed the mean of the weighted adjacency matrices for each diagnostic group and dataset, built the corresponding group-average Laplacian, and inspected its gap spectrum to determine \(k\). This \(k\) was then used to retain, for each subject, the first \(k\) eigenvectors (an \(n\times k\) orthonormal matrix) associated with the smallest eigenvalues. This step serves as a structured, graph-spectral preprocessing rather than a conventional feature-selection procedure.

Finally, we treated each subject’s eigenvector matrix as a point on the Grassmannian manifold and implemented a discriminant analysis by optimization on the Grassmannian. The goal was to estimate two subspace centers (one per diagnostic group) that maximize between-group separation while minimizing within-group dispersion.

\medskip
\noindent\textbf{Principal angles and Grassmannian distance.}  
For two orthonormal bases \(U,V\), the principal angles \(\{\theta_\ell\}_{\ell=1}^k\) between the subspaces they span are obtained from the singular values of \(M=U^\top V\). If \(M=Q\Sigma R^\top\) is the (compact) singular value decomposition and \(\Sigma=\mathrm{diag}(\sigma_1,\dots,\sigma_k)\) with \(\sigma_\ell\in[0,1]\), then
\[
\theta_\ell \;=\; \arccos(\sigma_\ell), \qquad \ell=1,\dots,k.
\]
The squared Grassmannian distance between the subspaces spanned by \(U\) and \(V\) is defined as the sum of squared principal angles:
\[
d_{\mathcal{G}}^2(U,V) \;=\; \sum_{\ell=1}^k \theta_\ell^2.
\]
This distance is invariant under right-multiplication of \(U\) or \(V\) by $k\times k$ orthogonal matrices and therefore measures the intrinsic discrepancy between subspaces rather than between particular basis representations.

\medskip
\noindent\textbf{Initialization: class centers as Fréchet means.}  
For each class \(c\in\{\mathrm{HC},\mathrm{PT}\}\) we initialize a class center \(C^{(c)}\) as the Fr\'echet mean on the Grassmannian, i.e., as the minimizer of the average squared Grassmannian distance:
\begin{equation}\label{eq:frechet-centers}
C^{(c)} \;=\; \arg\min_{C\in\mathrm{Gr}(k,n)}\frac{1}{N_c}\sum_{i\in\mathcal{I}_c} d_{\mathcal{G}}^2\big(C,\,U_i^{(c)}\big).
\end{equation}
In practice we compute \(C^{(c)}\) with a Karcher flow: starting from an initial guess (e.g., one sample or the Euclidean average followed by orthonormalization), iterate the intrinsic update \(C\leftarrow\Exp_{C}\big(\frac{1}{N_c}\sum_i\Log_{C}(U_i)\big)\) until convergence. Equation \eqref{eq:frechet-centers} yields the initial centroids \(C^{(\mathrm{HC})}_0\) and \(C^{(\mathrm{PT})}_0\) used by the optimizer.

\medskip
\noindent\textbf{Objective (cost) function: between- and within-domain terms.}  
The objective function is constructed to maximize interclass separation while simultaneously minimizing within-class dispersion. Denote the within-class dispersion by
\[
D_W(C^{(\mathrm{HC})},C^{(\mathrm{PT})}) \;=\; \sum_{i\in\mathcal{I}_{\mathrm{HC}}} d_{\mathcal{G}}^2\big(U_i^{(\mathrm{HC})},C^{(\mathrm{HC})}\big)
\;+\; \sum_{j\in\mathcal{I}_{\mathrm{PT}}} d_{\mathcal{G}}^2\big(U_j^{(\mathrm{PT})},C^{(\mathrm{PT})}\big),
\]
and the between-class separation by the squared distance between centers:
\[
D_B(C^{(\mathrm{HC})},C^{(\mathrm{PT})}) \;=\; d_{\mathcal{G}}^2\big(C^{(\mathrm{HC})},C^{(\mathrm{PT})}\big).
\]
The scalar objective to be maximized is the ratio \(D_B/(D_W+\varepsilon)\), where \(\varepsilon>0\) is a small numerical constant added to avoid division by zero in degenerate cases. Equivalently, we minimize the negative ratio:
\[
J(C^{(\mathrm{HC})},C^{(\mathrm{PT})}) \;=\; -\,\frac{D_B(C^{(\mathrm{HC})},C^{(\mathrm{PT})})}{D_W(C^{(\mathrm{HC})},C^{(\mathrm{PT})}) + \varepsilon}.
\]
This objective is a manifold generalization of the classical Fisher criterion (between-class variance over within-class variance), adapted to the Grassmannian geometry by replacing Euclidean variances with sums of squared geodesic distances. Convergence yields the optimized centers \(\big(C^{(\mathrm{HC})}_\star, C^{(\mathrm{PT})}_\star\big)\), which (locally) maximize interclass separation relative to intraclass dispersion.

\medskip
\noindent\textbf{Classification rule for novel samples.}  
Given a new observation \(U_{\mathrm{new}}\), its label is assigned by nearest-center classification with respect to the Grassmannian distance:
\[
\text{label}(U_{\mathrm{new}}) \;=\; 
\begin{cases}
\mathrm{HC}, & d_{\mathcal{G}}^2\big(U_{\mathrm{new}},C^{(\mathrm{HC})}_\star\big) \;<\; d_{\mathcal{G}}^2\big(U_{\mathrm{new}},C^{(\mathrm{PT})}_\star\big),\\[4pt]
\mathrm{PT}, & \text{otherwise.}
\end{cases}
\]
Thus, classification reduces to comparing squared principal-angle distances from \(U_{\mathrm{new}}\) to the two optimized centers. The use of principal angles as the local dissimilarity measure makes the approach invariant to orthonormal basis rotations and sign indeterminacies. 

We evaluated the classifier using 5-fold stratified nested cross-validation. In each outer fold, data were split into training and test sets. Class-specific centers were estimated from training samples of each class (controls and patients), yielding representative subspaces. Each test sample was then classified to the nearest class center. Performance metrics included accuracy, sensitivity, specificity, and AUC-ROC. This model was further compared to a linear discriminant analysis (LDA) classifier on the same input features using the same training protocol.


\section{Results}

\subsection{Hypothesis testing on correlation matrices}
\begin{figure}[!t] 
  \centering
  \includegraphics[width=0.85\textwidth]{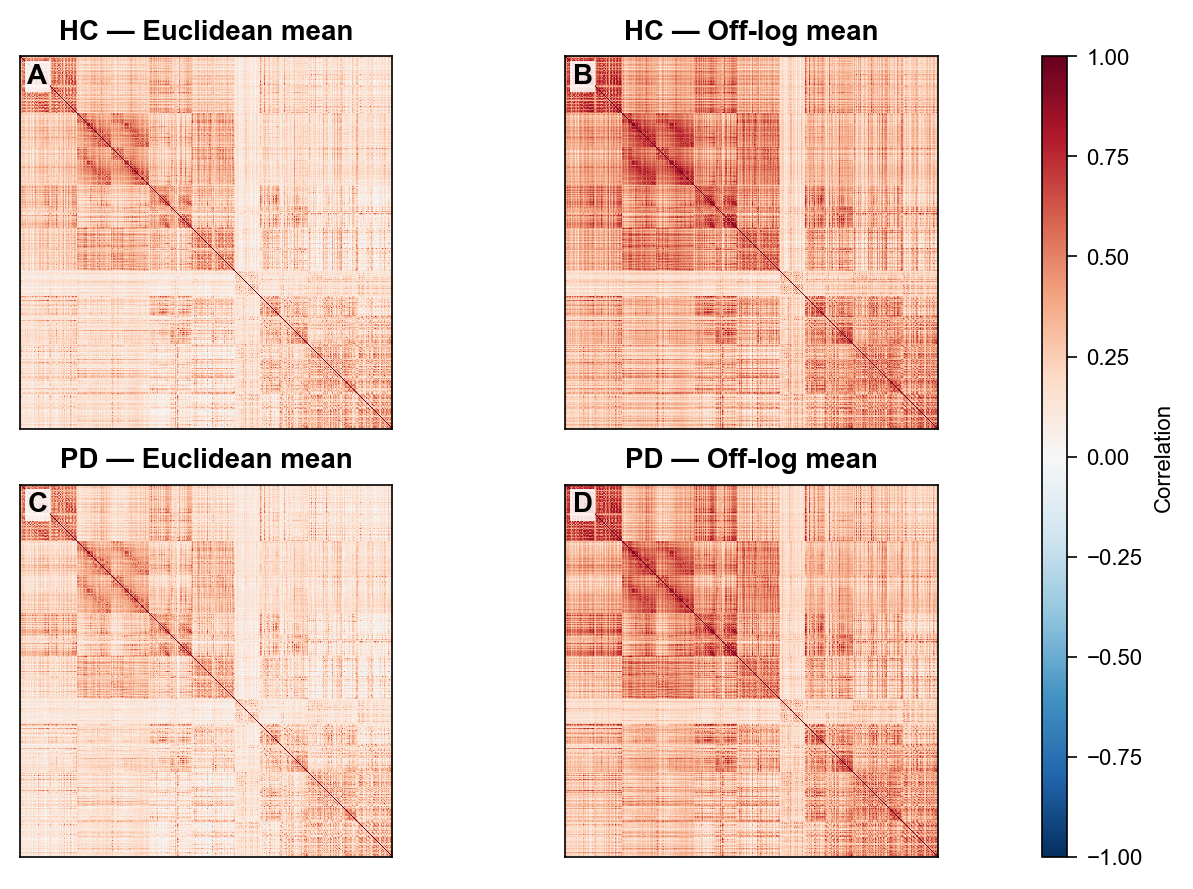} 
  \caption{Group-mean functional connectivity matrices (HC vs PD).(A–B) Healthy controls (HC); (C–D) Parkinson's disease (PD). Left panels show the element-wise Euclidean mean; right panels show the Fréchet mean on the correlation manifold computed with the Off--log metric.}
  \label{fig:fmri}
\end{figure}

\begin{figure}[!t] 
  \centering
  \includegraphics[width=0.85\textwidth]{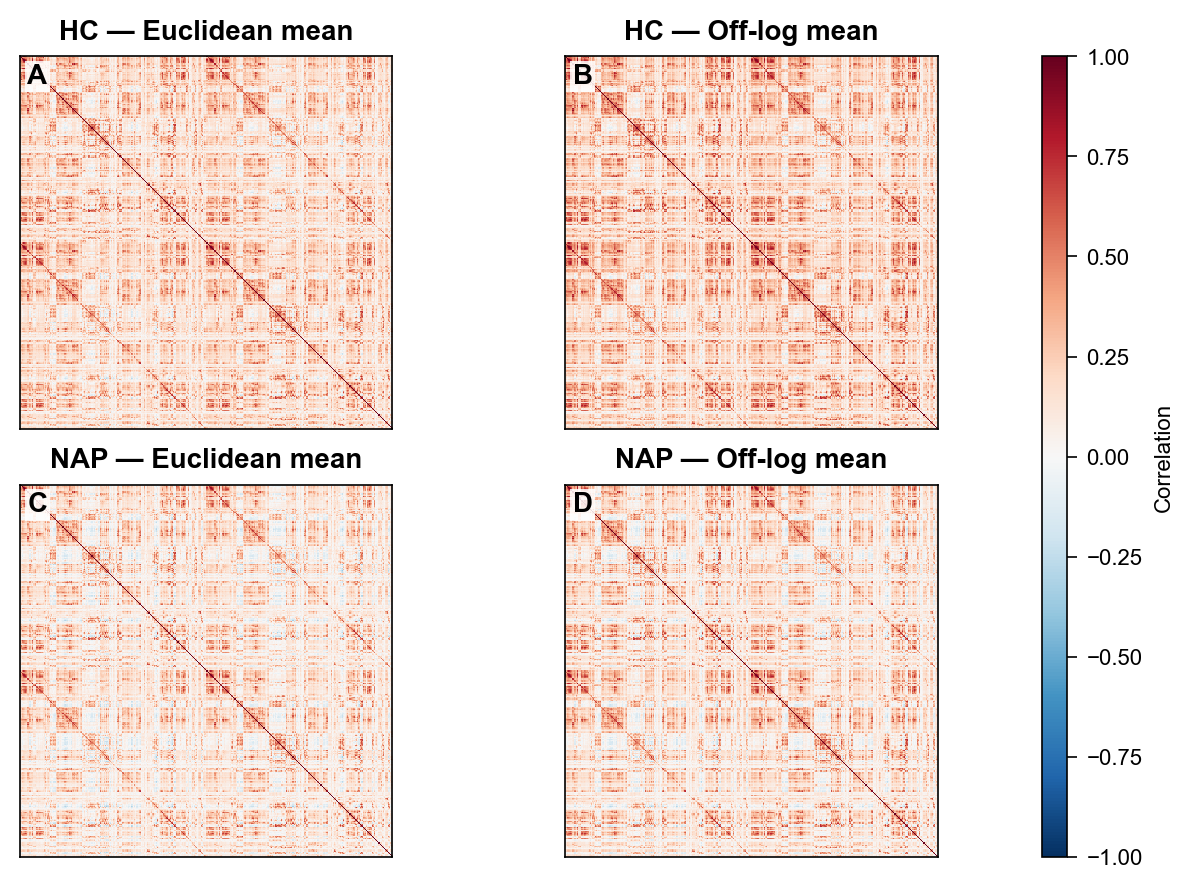} 
  \caption{Group-mean functional connectivity matrices (HC vs NAP).(A–B) Healthy controls (HC); (C–D) Non-affective psychosis (NAP). Left panels show the element-wise Euclidean mean; right panels show the Fréchet mean on the correlation manifold computed with the Off--log metric.}
  \label{fig:fmri_15}
\end{figure}

\begin{figure}[!t] 
  \centering
  \includegraphics[width=0.85\textwidth]{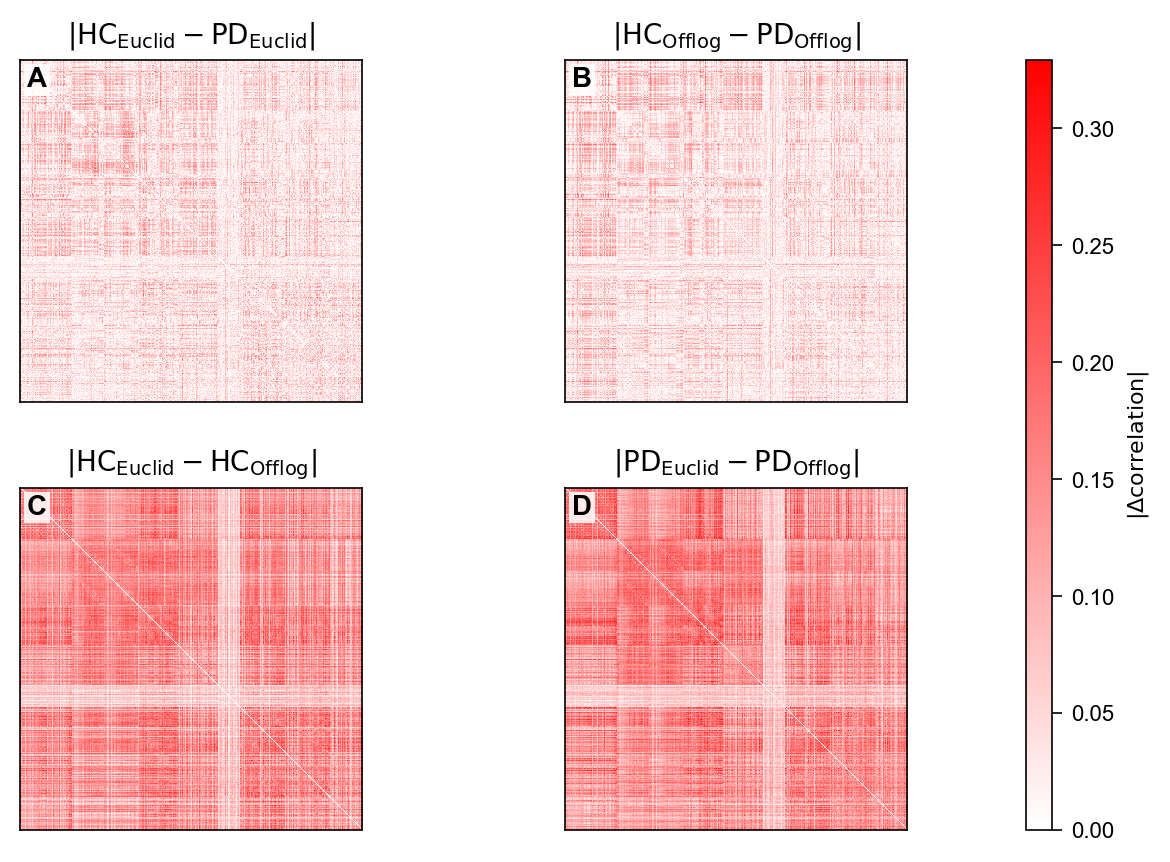} 
  \caption{Group-mean functional connectivity matrices difference (HC vs PD).(A–B) Healthy controls (HC) vs Parkinson's disease (PD) using both raw correlation matrices and Off--log transformed; (C–D) Within group differences using raw correlation matrices and Off--log transformed.}
  \label{fig:fmri_2}
\end{figure}

Figure~\ref{fig:fmri} and Figure~\ref{fig:fmri_15} illustrate the qualitative differences between the Euclidean mean and the Fréchet mean, computed under the Off--log metric, for both clinical groups. The difference is more pronounced for the \textit{DataPD} cohort, where the two estimates appear visually different, with the Fréchet mean exhibiting a more well-defined spatial pattern across the seven functional networks. Moreover, Figure~\ref{fig:fmri_2} reports the differences of the two means between the two groups of \textit{DataPD} cohort, as well as the differences within each group when using the two different geometric metrics.

Hypothesis testing was conducted using the Biswas--Ghosh permutation test (1,000 label permutations), applied to (i) raw correlation matrices and (ii) tangent-space-projected matrices obtained via the Off--log mapping. The results varied across datasets. For \textit{DataPD} (PD vs.\ HC), the Biswas--Ghosh statistic rejected the null hypothesis only when tangent-space-projected data were used ($p$-value = 0.01), whereas the same test applied to raw correlation matrices did not reach statistical significance ($p$-value = 0.37). In contrast, for \textit{DataNAP} (HC vs.\ NAP), we observed statistically significant group separation for both raw correlation matrices and tangent-space representations ($p$-value $<$ 0.001). These findings suggest that, in our PD cohort, the Off--log tangent projection may enhance sensitivity to disease-related connectivity differences that remain undetected under a Euclidean analysis. In \textit{DataNAP}, however, the difference appears sufficiently large to be captured by both representations.

\subsection{Brain age prediction}
We evaluated brain-age regression on the three ageing cohorts (\textit{cam-CAN, HCPAging and NKI}) of HC using an Elastic Net regressor trained on two types of inputs: (i) raw correlation matrices and (ii) tangent-space projections (ECM and Off--log representations). Performance was assessed with MAE and $R^2$ averaged across cross-validation folds; results are summarized in Table~\ref{tab:brainage}.

Overall, all geometry-aware representations achieved competitive age-prediction performance. In our experiments, raw correlation matrices yielded better MAE and $R^2$ values than the geometric alternatives only in the \textit{cam-CAN} cohort, but both raw and Off--log representations achieved state-of-the-art performance on this dataset. In the other two datasets, the ECM metric showed superior performance than both raw correlations and the Off--log metric (Table~\ref{tab:brainage}). The key finding is that mapping to a geometry-aware representation does not compromise predictive accuracy and provides a principled, constraint-respecting framework for downstream statistical analysis, performing at least as well as raw correlations and, in some cases, even better. 

\begin{table}[ht]
\centering
\caption{Brain-age prediction. Reported metrics are cross-validated MAE and $R^2$. Best values are in \textbf{bold}.}
\label{tab:brainage}
\begin{tabular}{lcc}
\hline
Input representation & MAE (years) & $R^2$ \\
\hline
\multicolumn{3}{c}{cam-CAN dataset}\\
\hline
\textbf{Raw correlation} & $\bm{6.782 \pm 0.354}$ & $\bm{0.783 \pm 0.032}$ \\
Off--log & $6.968 \pm 0.393$ & $0.768 \pm 0.038$ \\
ECM & $7.170 \pm 1.165$ & $0.698 \pm 0.175$ \\
\hline
\multicolumn{3}{c}{HCP-Aging dataset}\\
\hline
Raw correlation & $5.539 \pm 0.304$ & $0.776 \pm 0.032$ \\
Off--log & $5.516 \pm 0.474$ & $0.778 \pm 0.045$ \\
\textbf{ECM} & $\bm{5.280 \pm 0.371}$ & $\bm{0.794 \pm 0.018}$ \\
\hline
\multicolumn{3}{c}{NKI dataset}\\
\hline
Raw correlation & $6.622 \pm 0.639$ & $0.778 \pm 0.040$ \\
Off--log & $6.845 \pm 0.411$ & $0.766 \pm 0.029$ \\
\textbf{ECM} & $\bm{6.380 \pm 0.409}$ & $\bm{0.787 \pm 0.039}$ \\
\hline
\end{tabular}
\end{table}

\subsection{Classification}

Classification experiments were conducted on the same two clinical datasets (\textit{DataPD}: HC vs.\ PD; \textit{DataNAP}: HC vs.\ NAP). Table~\ref{tab:classification} summarizes the cross-validated performance obtained using four input representations: raw correlation, Off--log, ECM, and LEC.
Across both datasets, the Off--log tangent representation achieved the best classification performance on all reported metrics. These improvements were consistent for accuracy and AUC-ROC and were also reflected in sensitivity and specificity. Overall, these findings suggest that the Off--log geometry provides features that may be more discriminative for the diagnostic distinctions examined.

\begin{table}[ht]
\centering
\caption{Cross-validated classification performance. For each dataset we compared raw, Off--log, ECM, and LEC inputs. Best values are highlighted in \textbf{bold}.}
\label{tab:classification}
\begin{tabular}{lcccc}
\hline
Input representation & Accuracy & AUC-ROC & Sensitivity & Specificity \\
\hline
\multicolumn{5}{c}{DataPD (HC vs.\ PD)}\\
\hline
Raw     & $0.461 \pm 0.145$ & $0.517 \pm 0.172$ & $0.500 \pm 0.354$ & $0.450 \pm 0.277$ \\
Off--log& $\bm{0.693 \pm 0.174}$ & $\bm{0.762 \pm 0.183}$ & $\bm{0.700 \pm 0.245}$ & $\bm{0.700 \pm 0.215}$ \\
ECM     & $0.532 \pm 0.196$ & $0.592 \pm 0.201$ & $\bm{0.700 \pm 0.187}$ & $0.333 \pm 0.365$ \\
LEC     & $0.536 \pm 0.175$ & $0.567 \pm 0.233$ & $0.650 \pm 0.200$ & $0.400 \pm 0.226$ \\
\hline
\multicolumn{5}{c}{DataNAP (HC vs.\ NAP)}\\
\hline
Raw     & $0.657 \pm 0.110$ & $0.766 \pm 0.104$ & $0.717 \pm 0.145$ & $0.592 \pm 0.119$ \\
Off--log& $\bm{0.701 \pm 0.084}$ & $\bm{0.812 \pm 0.075}$ & $0.667 \pm 0.175$ & $\bm{0.739 \pm 0.106}$ \\
ECM     & $0.699 \pm 0.122$ & $0.741 \pm 0.126$ & $0.717 \pm 0.085$ & $0.679 \pm 0.163$ \\
LEC     & $0.683 \pm 0.125$ & $0.772 \pm 0.112$ & $\bm{0.767 \pm 0.062}$ & $0.594 \pm 0.211$ \\
\hline
\end{tabular}
\end{table}

\subsection{Grassmannian discriminant analysis}

\begin{figure}[!t] 
  \centering
  \includegraphics[width=0.9\textwidth]{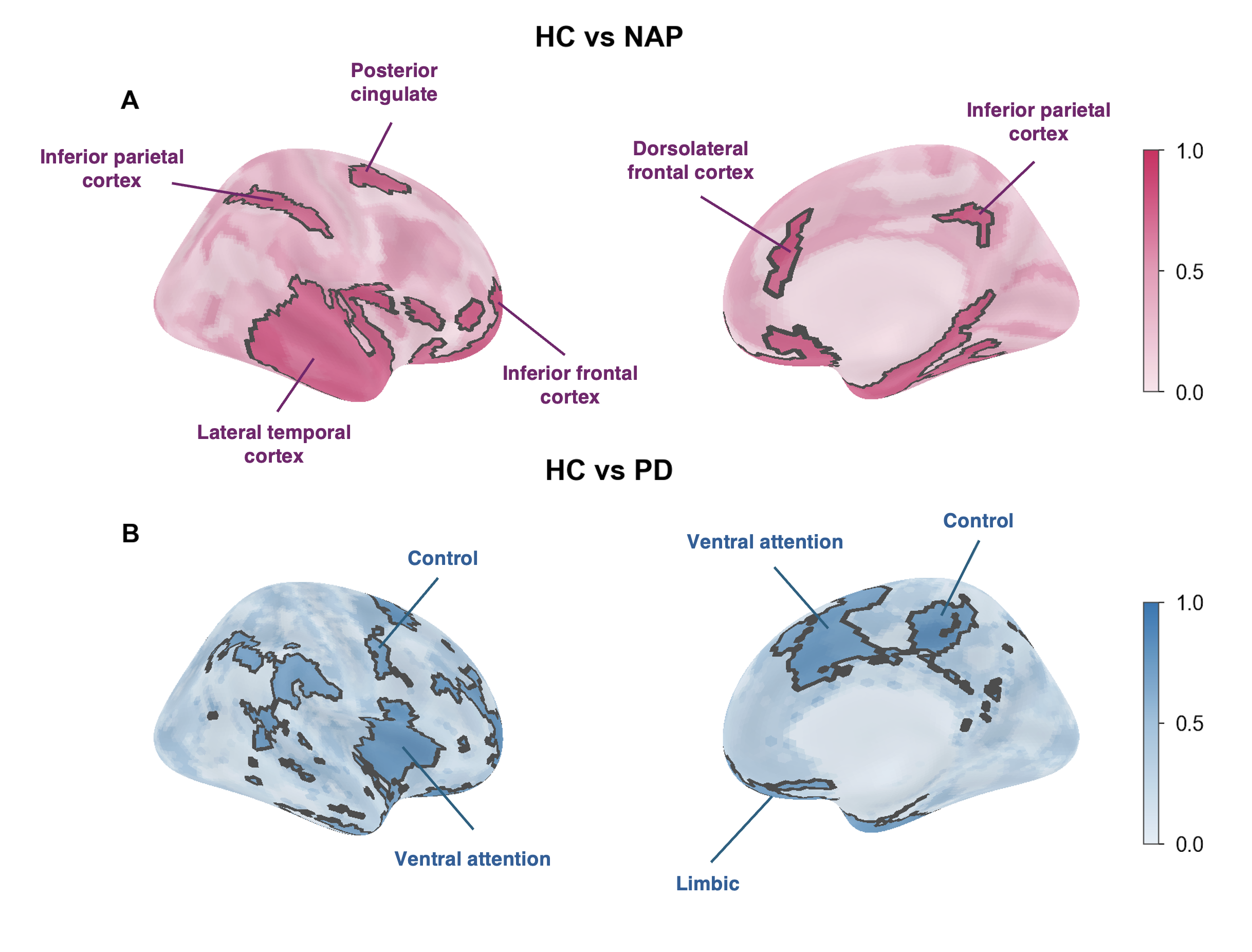} 
  \caption{Cortical regions most predictive of group membership identified by Grassmannian discriminant analysis. (A) HC vs NAP; (B) HC vs PD. Panels show lateral views of the left and right hemispheres. Colored parcels mark areas with the largest absolute discriminant weights for the indicated contrast (pink: HC vs NAP; blue: HC vs PD). Marked regions correspond to the parcels most consistently selected across the 5 cross-validation folds (i.e., those with the highest selection frequency).}
  \label{fig:brain}
\end{figure}

The Grassmannian discriminant pipeline was evaluated on the same two clinical datasets and compared against a baseline LDA classifier (Table~\ref{tab:grassmann}). Across both datasets, the Grassmannian discriminant consistently outperformed LDA, which did not surpass chance-level performance, with improvements that were robust across cross-validation folds. For each fold, we also identified the regions contributing most strongly to discrimination between groups.
In the HC vs.\ PD task, the majority of discriminative regions were located within the salience and limbic functional networks, with a smaller number of regions associated with the control network (Figure~\ref{fig:brain}). In contrast, for the HC vs.\ NAP task, the most discriminative regions included the inferior parietal cortex, posterior cingulate, lateral temporal cortex, inferior frontal cortex, and dorsolateral prefrontal cortex---all areas consistently implicated in psychosis-related dysfunctions \cite{Li2017-kh, Fu2021-uk} (Figure~\ref{fig:brain}).
These results suggest that representing eigenspaces as points on the Grassmannian manifold and performing discrimination directly in subspace provides more stable and informative features than treating eigenvectors as ordinary Euclidean vectors, which are susceptible to sign and basis indeterminacies. Moreover, this approach enabled the identification of discriminant patterns that align closely with known disease-specific neural signatures \cite{Li2017-kh, Fu2021-uk, Tessitore2017-ae, Liu2022-sm}, highlighting its potential utility in uncovering meaningful biomarkers from functional connectivity data.

\begin{table}[ht]
\centering
\caption{Cross-validated classification performance. For each dataset we compared Grassmannian and Euclidean representations. Best values are in \textbf{bold}.}
\label{tab:grassmann}
\begin{tabular}{lcccc}
\hline
Input representation & Accuracy & AUC-ROC & Sensitivity & Specificity \\
\hline
\multicolumn{5}{c}{Dataset1 (HC vs.\ PD)}\\
\hline
Euclidean     & $0.543 \pm 0.035$ & $0.558 \pm 0.154$ & {1.000 $\pm 0.000$} & $0.000 \pm 0.000$ \\
Grassmannian  & $\bm{0.693 \pm 0.216}$ & $\bm{0.567 \pm 0.146}$ & $\bm{0.700 \pm 0.292}$ & $\bm{0.700 \pm 0.215}$ \\
\hline
\multicolumn{5}{c}{Dataset2 (HC vs.\ NAP)}\\
\hline
Euclidean     & $0.463 \pm 0.089$ & $0.454 \pm 0.058$ & $0.583 \pm 0.091$ & $0.335 \pm 0.175$ \\
Grassmannian  & $\bm{0.692 \pm 0.089}$ & $\bm{0.616 \pm 0.154}$ & $\bm{0.767 \pm 0.111}$ & $\bm{0.612 \pm 0.112}$ \\
\hline
\end{tabular}
\end{table}

\paragraph{Summary of results.}
The Off--log Euclidean geometry offers a practical and effective representation of fMRI correlation matrices for both exploratory statistical analyses and supervised learning tasks. In the exploratory setting, it increased sensitivity in permutation-based two-sample testing (\textit{DataPD}), while in downstream supervised tasks it delivered improved classification performance compared to raw, ECM, and LEC inputs. For brain-age regression, raw correlation matrices yielded slightly better point estimates in \textit{cam-CAN} cohort; however, ECM representations achieved better performances on \textit{HCPAging} and \textit{NKI} datasets. Finally, applying Grassmannian methods to Laplacian eigenvector subspaces provided clear advantages over conventional Euclidean treatments of eigenvectors, yielding more stable and clinically informative features.


\section{Discussion}

We evaluated the Off--log Euclidean geometry as a practical representation for fMRI correlation matrices and compared it to several alternatives (Euclidean, ECM, and LEC) across a set of complementary tasks: exploratory summary statistics and two-sample testing, brain-age regression, diagnostic classification, and subspace discrimination. Below we discuss the principal findings.

\subsection{Interpretation and methodological implications}

The results highlight two recurring themes. The first is \emph{geometry matters}: conducting statistical analysis or machine learning in coordinates that respect the manifold structure of correlation matrices can change sensitivity, effect-size estimation, and discriminative power. In particular, the Off--log representation provides a principled coordinate system (a Lie-algebra parametrization) that is permutation invariant and metric consistent; operations on the manifold reduce to Euclidean operations in this coordinate system and map back to valid correlation matrices via \(\Expoff\). These observations align with a growing literature showing that geometry-aware formulations improve performance on neuroimaging tasks \cite{Pennec2019-up, Ju2025-rj}. Across modalities and problem settings, non-Euclidean treatments of matrix-valued signals have demonstrated clear advantages in both theory and practice. Foundational work established Riemannian frameworks for SPD tensors and diffusion MRI \cite{Fletcher2007-vp}, together with well-posed metrics (e.g., log-Euclidean and affine-invariant) and practical algorithms \cite{Fillard2007-jv}. In electrophysiology, methods that operate directly on covariance structure have yielded strong results for EEG/Brain computer interface decoding and feature extraction \cite{Barachant2012-pv, Tibermacine2024-tk}, motivating analogous treatments of functional connectivity in fMRI \cite{You2022-vj, You2025-ta}. Building on this foundation, recent studies have applied Riemannian tools to functional connectivity matrices for multisite harmonization, longitudinal modeling, and time-varying connectivity, with tangible gains in inference quality and robustness to site/batch effects \cite{Honnorat2024-kv}. Parallel developments extend these ideas to generative modeling and flow-based sampling on matrix manifolds via pullback geometries \cite{Collas2025-ij}. Taken together, these advances substantiate our findings and reinforce the view that geometry-aware methods offer a principled and effective approach to matrix-valued neuroimaging data.

The second theme is \emph{task dependency} of the best geometric representation. In our experiments, brain-age regression did not improve when using Off--log instead of raw correlations, whereas the ECM metric yielded slightly better performance in two of the three cohorts examined. By contrast, binary diagnostic classification benefited consistently from Off--log projection. Thus, while Off--log appears advantageous for detecting subtle network-pattern differences and for classification, it is not universally superior for all supervised regression settings; task characteristics and the origin of signal determine which representation is most effective.

More broadly, our findings underscore a fundamental methodological distinction between \emph{tangent-space} and \emph{manifold-native} approaches. Tangent-space approaches map data to the tangent space at a base point on the manifold, where the curved geometry locally flattens. This linearization dramatically simplifies computations: distances, means, and linear models can be applied as if operating in Euclidean space, while still approximately respecting the underlying geometry. However, this simplification comes at a cost—the tangent representation is only locally accurate around the base point and may distort global structure when the data are widely dispersed. Thus, the choice of the most appropriate base point is non-trivial and may depend on the type of dataset. Notably, for the Off--log metric, the projection at the identity coincides with the relevant Lie algebra and, under appropriate conditions, defines a global diffeomorphism \cite{Bisson-undated-nh}.

In contrast, \emph{manifold-native} methods operate directly on the curved manifold without flattening. Our Grassmannian discriminant analysis exemplifies this: the algorithm remained entirely within the Grassmannian manifold and preserved its intrinsic geometry throughout. This native treatment avoids the approximation errors inherent in tangent projections, leading to improved performance in our experiments. However, it is also computationally more demanding, requiring specialized solvers and matrix operations that respect the manifold constraints at every step. 

These observations imply an important methodological trade-off: tangent projections are easier and computationally cheaper but approximate, whereas manifold-native methods are geometrically faithful and often more accurate, yet heavier. Moreover, tangentization does not guarantee improved downstream performance, it can either help or hinder, depending on how well the linearized space aligns with the structure of the signal. Still, knowing the underlying geometry is valuable: even if one ultimately applies a tangent-based approach for practical reasons, understanding where and why it approximates the manifold can guide the choice of base point, inform interpretation, and inspire hybrid designs that mix local linearization with occasional reprojection to the manifold to reduce drift.

\subsection{Limitations}

Several constraints of the present study should be acknowledged. First, the datasets provide a useful but non-exhaustive view of clinically relevant heterogeneity; results may differ under alternative cohorts, parcellations, preprocessing pipelines, or tasks. Sample sizes for the classification tasks are modest by design, to reflect realistic experimental medicine settings, but this limits statistical power. Because connectivity structure and distributional properties can vary across acquisition protocols, preprocessing strategies, and atlas resolutions, the robustness of the observed Off--log advantages under such shifts remains to be established. Larger, multisite evaluations will be necessary to assess generalizability and stability.

Second, the classification and regression baselines were intentionally simple (linear SVM and Elastic Net) to isolate representational effects; more powerful nonlinear models (e.g., kernel SVMs, graph neural networks, or deep manifold networks) could either attenuate or magnify the influence of the chosen geometry. While this choice allows clearer attribution of observed effects to the representation itself, it also limits the practical performance ceiling. The interaction between geometric representation and model capacity remains an open question and could lead to different conclusions in more expressive architectures.

Third, the employed Riemannian metrics focus on full-rank correlation matrices: when empirical matrices are rank deficient (e.g., due to short time series or high-dimensional parcellations), careful regularization or shrinkage is necessary before applying the matrix logarithm and exponential operations. Furthermore, when data are widely dispersed across the manifold, the tangent-space approximation can degrade, potentially underestimating distances and biasing statistical estimates. In such cases, native manifold approaches may be preferable despite their higher computational burden.

\section{Conclusion}
\label{sec:conclusion}

In this work, we showed that treating fMRI correlation matrices as points on appropriate manifolds, rather than as unconstrained Euclidean vectors, leads to measurable gains in sensitivity, stability, and downstream predictive performance, while being supported by a rigorous and precise mathematical theory. On the correlation side, the Off--log diffeomorphism provides a permutation-invariant, log–Euclidean–like geometry, yielding closed-form distances, Fréchet means, and tangent-space linearization while mapping back to valid correlations. This, in turn, makes it possible to deploy standard statistical learning pipelines on geometrically corrected data without violating the correlation structure.
We further argued that whenever the information is fundamentally spectral, as with graph-Laplacian eigenvectors, the natural object is the subspace, not the ordered basis. Modeling these eigenspaces on the Grassmannian, and performing discrimination directly in subspace space, removes sign/basis ambiguities and produces more stable and interpretable patterns than Euclidean baselines, highlighting disease-relevant networks in both cohorts. Taken together, these two components constitute a coherent and scalable geometric toolkit that can be readily integrated into existing neuroimaging workflows for hypothesis testing, classification, and regression.

\appendix
\section{fMRI data preprocessing}
\label{sec:appendix}

Preprocessing of anatomical and functional data of the \textit{cam-CAN}, \textit{HCPAging}, \textit{NKI} and the \textit{DataPD} datasets was performed using fMRIPrep version 23.2.2 \cite{Esteban2019-mc}. The T1w image was corrected for intensity non-uniformity and skull-stripped. Brain surfaces were reconstructed using ‘recon-all’ from FreeSurfer \cite{Fischl2012-wr}, and the brain mask estimated previously was refined with a custom variation of the method to reconcile ANTs-derived and FreeSurfer-derived segmentation of the cortical gray matter (GM). Spatial normalization to the ICBM 152 Nonlinear Asymmetrical template version 2006 was performed through nonlinear registration, using brain-extracted versions of both the T1w volume and the template. Brain tissue segmentation of cerebrospinal fluid, white matter (WM), and GM was performed on the brain-extracted T1w using FSL's FAST tool. Only for \textit{DataPD}, the first 8 volumes of the rs-fMRI acquisition were removed, to allow the magnetization to reach the steady state. Functional data was then slice-time corrected and motion corrected. This was followed by co-registration to the corresponding T1w using boundary-based registration with six degrees of freedom. Motion correction transformations, BOLD-to-T1w transformation, and T1w-to-template (MNI) warp were concatenated and applied in a single step. Physiological noise regressors were extracted using CompCor. Principal components were estimated using aCompCor. A mask to exclude signal with cortical origin was obtained by eroding the brain mask, ensuring it only contained subcortical structures. For aCompCor, six components were calculated within the intersection of the subcortical mask and the union of cerebrospinal fluid and white matter masks calculated in T1w space, after their projection to the native space of each functional run. Frame-wise displacement (FD) was calculated for each functional volume.
The fMRIPrep functional outputs were then subjected to additional processing using XCP-D version 0.7.4 \cite{Mehta2023-rm}. After interpolating high-motion outlier volumes (FD $>$ 0.5 mm) with cubic spline, confound regression was applied to the fMRI data. The regression matrix included six motion parameters and their derivatives, five aCompCor components relative to CSF and five to WM. Finally, high-pass temporal filtering with a cut-off frequency of 0.001 Hz was applied, followed by spatial smoothing using a Gaussian kernel with a FWHM of 6 mm. Details on the preprocessing of \textit{DatasetNAP} have been published previously and can be found in the following references \cite{Hancock2023-eg, Shenton2019-va}.

\section*{Data availability}
The data and code supporting the findings of this study are available from the corresponding author upon reasonable request.

\printbibliography

\section*{Funding}
This research is funded by the Ministry of University and Research within the Complementary National Plan PNC-I.1 "Research initiatives for innovative technologies and pathways in the health and welfare sector, D.D. 931 of 06/06/2022, PNC0000002 DARE - Digital Lifelong Prevention CUP: B53C22006440001.

\section*{Acknowledgements}
MV is supported by EU funding within the MUR PNRR “National Center for HPC, BIG DATA AND QUANTUM COMPUTING (Project no. CN00000013 CN1), the Ministry of University and Research within the Complementary National Plan PNC DIGITAL LIFELONG PREVENTION - DARE (Project no PNC0000002\_DARE), by Fondo per il Programma Nazionale di Ricerca e Progetti di Rilevante Interesse Nazionale (PRIN), (Project no 2022RXM3H7), and by by the cascading grant “Q Amyloid – Quantitative Amyloid Imaging” under PNRR ECS00000017 “THE - Tuscany Health Ecosystem,” Spoke 6: “Precision Medicine \& Personalized Healthcare,” funded by the European Commission under the NextGeneration EU programme. We acknowledge the use of data from three public databases. Data collection and sharing for this project were provided by the Cambridge Centre for Ageing and Neuroscience. Cambridge Centre for Ageing and Neuroscience funding was supported by the UK Biotechnology and Biological Sciences Research Council (grant number BB/H008217/1), together with support from the UK Medical Research Council and the University of Cambridge, UK. Additionally, data from the HCP-Aging 2.0 Release were utilized in this research. Research reported in this publication was supported by the National Institute on Aging of the National Institutes of Health under Award Number U01AG052564 and by funds provided by the McDonnell Center for Systems Neuroscience at Washington University in St. Louis. The Lifespan Human Connectome Project in Aging 2.0 Release data used in this report came from DOI: 10.15154/1520707. We also acknowledge the Enhanced Nathan Kline Institute–Rockland Sample for providing open-access neuroimaging data; data collection and sharing for the Enhanced Nathan Kline Institute–Rockland Sample were supported by National.

\section*{Author contributions}

M.S. conceived the study, developed the theoretical framework, designed the methodology, and performed the experiments. M.V. and M.M. supervised the project and contributed to the interpretation and validation of the results. M.S. drafted the manuscript, and all authors critically revised it for important intellectual content. All authors read and approved the final version of the manuscript.

\section*{Declarations}

\section*{Competing interests}
The authors declare no competing interests.

\end{document}